\documentclass[lettersize,journal]{IEEEtran}
\usepackage{amsmath,amsfonts}
\usepackage{algorithmic}
\usepackage{array}
\usepackage[caption=false,font=normalsize,labelfont=sf,textfont=sf]{subfig}
\usepackage{textcomp}
\usepackage{stfloats}
\usepackage{url}
\usepackage{bm}
\usepackage{verbatim}
\usepackage{graphicx}
\usepackage{multirow}
\usepackage[table,xcdraw]{xcolor}
\usepackage{float}
\usepackage{graphicx}
\usepackage{subfig}
\usepackage{epstopdf}
\usepackage{color}
\usepackage{makecell}
\usepackage[caption=false,font=normalsize,labelfont=sf,textfont=sf]{subfig}
\usepackage{mathrsfs}
\usepackage{xcolor}
\usepackage{array}
\newcolumntype{C}[1]{>{\centering\arraybackslash}p{#1}}

\usepackage{float}
\usepackage{multirow}
\usepackage{booktabs}

\usepackage{pifont}
\usepackage{amssymb}

\usepackage[numbers,sort&compress]{natbib} 
\usepackage{color} 
\usepackage{hyperref}
\hypersetup{hypertex=true,
            colorlinks=true,
            linkcolor=blue,
            anchorcolor=blue,
            citecolor=blue}

\begin{document}
\title{Texture-guided Coding for Deep Features}
\author{Lei Xiong, Xin Luo, Zihao Wang, Chaofan He, Shuyuan Zhu,~\IEEEmembership{Member,~IEEE}, and Bing Zeng,~\IEEEmembership{Fellow,~IEEE}
\thanks{Lei Xiong, Xin Luo, Zihao Wang, Shuyuan Zhu, and Bing Zeng are with the School of Information and Communication Engineering, University of Electronic Science and Technology of China, Chengdu 611731, China.}
\thanks{Chaofan He is with the Peng Cheng Laboratory, Shenzhen 518066, China.}}

\maketitle
	
	\begin{abstract}
		
	With the rapid development of machine vision technology in recent years, many researchers have begun to focus on feature compression that is better suited for machine vision tasks. The target of feature compression is deep features, which arise from convolution in the middle layer of a pre-trained convolutional neural network. However, due to the large volume of data and high level of abstraction of deep features, their application is primarily limited to machine-centric scenarios, which poses significant constraints in situations requiring human-computer interaction. This paper investigates features and textures and proposes a texture-guided feature compression strategy based on their characteristics. Specifically, the strategy comprises feature layers and texture layers. The feature layers serve the machine, including a feature selection module and a feature reconstruction network. With the assistance of texture images, they selectively compress and transmit channels relevant to visual tasks, reducing feature data while providing high-quality features for the machine. The texture layers primarily serve humans and consist of an image reconstruction network. This image reconstruction network leverages features and texture images to reconstruct preview images for humans. Our method fully exploits the characteristics of texture and features. It eliminates feature redundancy, reconstructs high-quality preview images for humans, and supports decision-making. The experimental results demonstrate excellent performance when employing our proposed method to compress the deep features.

	\end{abstract}
	
	\begin{IEEEkeywords}
		Deep feature compression, human–machine vision, machine vision
	\end{IEEEkeywords}
	
	\section{Introduction}

	\IEEEPARstart{I}{n} recent years, with the rapid development of machine vision technology, intelligent applications based on machine vision have been widely deployed in various fields. In these applications, machines must analyze and process massive amounts of image data, significantly challenging storage and transmission. Traditional image compression algorithms are mainly designed for human visual perception, aiming to minimize image distortion and achieve better visual quality. However, the requirements of machine vision for image compression are different from those of human visual perception. Machine vision mainly deals with features extracted from images, and the quality of features directly affects the performance of visual tasks. Existing image compression algorithms cannot satisfy the needs of machine vision. To address this issue, researchers have proposed a series of deep feature compression methods for machine vision in recent years. In this compression, the deep features are firstly extracted from the uncompressed image and output from the shallow network layers. Then, they are compressed and transmitted to the machine for vision tasks. However, how to compress intermediate deep features targeting high representation ability and low bit cost is still a challenging problem.

	The deep features are composed of several data channels, and each channel's data is like a video frame. Thus,  the classical video codec, such as high-efficiency video coding (HEVC)  \cite{hevc}, can compress deep features. For example,  in \cite{hevc_yolo},  the HEVC range extensions \cite{rext} were used to compress features extracted from YOLO \cite{yolo},  where the data of all the feature channels were combined to form a single image-like data. Different from \cite{hevc_yolo},  the data of each feature channel was processed as one video frame in  \cite{chen}  where the HEVC intra coding was employed to compress features extracted from the shallow layer of VGG \cite{vgg} and ResNet  \cite{resnet}. Although using the HEVC intra-coding reduces the intra-channel redundancy of features, inter-channel redundancy still exists between features. To address this problem, \cite{inter-feature} proposed to remove this redundancy by tiling all features into one frame and applying intra-prediction on the composed frame. Based on this work, \cite{spatio-temporal_arrangement} further divided the feature channels into groups according to the similarity and then tiled features of each group into a single frame for the compression with the HEVC inter-frame coding,  removing both intra-channel and inter-channel redundancies. 

    The video-based solution reduces the redundancy of features; however, it typically assigns the same quantization parameter (QP) to different feature channels. This oversight ignores the importance of the feature and limits the feature's ability to represent effectively. To address this issue, the sensitivity-aware bit allocation \cite{ref1} was introduced. This approach defines the sensitivity of each feature channel based on its impact on the fidelity of the computer vision task. By assigning different QPs to feature channels with varying sensitivities, adaptive bit allocation is achieved.

    Although the methods above effectively reduce the feature data, they still face some challenges. The feature maps obtained after a series of convolutions on images are highly abstract and can only be used by machines. Humans cannot directly understand their content, which has excellent limitations in scenarios that require human intervention. To address this issue, many researchers\cite{human-machine, face, scale_image_coding, sketch} have proposed compression methods for humans and machines. However, these methods do not fully utilize human and machine vision characteristics and lack research on the collaborative work of texture and features.

	 In this work, we propose a texture-guided feature compression method. It can simultaneously provide natural images and deep features for humans and machines, respectively, and fully utilizes texture images to remove feature redundancy. Specifically, we construct a two-layer encoding framework consisting of a feature layer and a texture layer. The feature layer mainly serves machine vision tasks and can also assist in image reconstruction of the texture layer; the texture layer serves humans, providing preview images for humans and reducing the amount of feature data transmitted. During feature encoding, we use a channel selection module to select task-related channels for transmission to reduce the amount of feature data. The missing channels are filled with texture images at the decoding end. In the texture image encoding stage, we downsample it to reduce the amount of data. At the decoding end, we use our proposed image reconstruction neural network to reconstruct the image by combining texture and features.

	The main contributions of this paper are as follows:
	\begin{enumerate}
		\item To the best of our knowledge, we are the first to attempt to use texture images to guide feature compression. We propose a texture-guided feature compression method that can fully utilize texture images to remove feature redundancy while improving the compression rate and maintaining the performance of visual tasks.
		
		\item We propose an image compression framework that can serve machines and humans. The framework consists of a feature layer and a texture layer. The feature layer mainly serves machine vision tasks, while the texture layer mainly serves humans.  

		\item We propose a feature reconstruction neural network that fully utilizes texture and features to achieve high-quality feature reconstruction.
		
		\item We propose an image reconstruction neural network that fuses features and textures. The network can fully utilize texture and features to complete image reconstruction.

	\end{enumerate}

 The rest of this paper is organized as follows. Section \ref{section:related_work} reviews some relevant works. Section \ref{section:proposed_method}  first introduces our framework for compression, followed by a description of the feature compression neural network serving machine vision tasks and the image reconstruction neural network serving humans. Section \ref{section:experiments} demonstrates the experimental results, and Section \ref{section:conclusion} concludes this paper.
	\section{Related Work}
        \label{section:related_work}

	\subsection{Deep Feature Compression with Codec} 
	In recent years, there has been an increasing amount of literature on deep feature compression.  Sufficiently compressing features without significantly degrading the performance of CV tasks is a common goal. In this subsection, we roughly divide the compression of depth features into lossless and lossy compression. 
	
	Lossless compression treats depth features as generic data and compresses them using GZIP\cite{gzip}, ZLIB\cite{zlib}, BZIP2\cite{bzip2} or LZMA\cite{lzma}. Although lossless compression achieves the best performance for the task, it has a limited compression ratio and does not adequately compress the deep feature data.

	In lossy compression, deep features are compressed by classical video codec,  such as high-efficiency video coding (HEVC)\cite{hevc}. In \cite{hevc_yolo}, the HEVC range extensions\cite{rext} were applied to compress the deep features which were extracted from YOLO \cite{yolo}, where all the channels were combined to form a single image \cite{hevc_yolo}. Different from \cite{hevc_yolo}, the feature of each channel was processed as the video frame in \cite{chen} where the HEVC intra\cite{hevc-intra} coding was used to compress all the feature channels extracted from the shallow layer of VGG \cite{vgg}, ResNet \cite{resnet}, etc. Deep features have similarities not only within each channel but also between channels. In Paper \cite{inter-feature}, the authors explored techniques for eliminating redundancy among feature channels. They tested three methods, including two inter-frame\cite{hevc-inter} approaches and one intra-frame\cite{hevc-intra} technique that tiled all channels into a single frame. The results showed that the tiling method performs best. Suzuki \emph{et al.}\cite{spatio-temporal_arrangement} also investigated methods for removing redundancy among feature channels. They rearrange all channels of a feature into two or three large feature maps and then compress them using HEVC inter-mode.

    The video-based solution reduces the redundancy of features. However, it usually assigns the same quantization parameter (QP) to different feature channels, which ignores the importance of the feature and cannot achieve a high representation ability for the feature. To solve this problem, the sensitivity-aware bit allocation was proposed\cite{ref1}, where the sensitivity of each feature channel was defined by its influence on the fidelity to the CV task. Moreover, different QP is assigned to the feature channel with different sensitivity, which achieves the adaptive bit allocation for features. Specifically, the feature channel with higher sensitivity would be allocated more bits to achieve high fidelity.
    In comparison, the channel with lower sensitivity would be allocated with fewer bits to guarantee that the overall bits do not exceed the given bit cost. In addition, Wang \emph{et al.}\cite{channel-wise_quantization} investigated the quantization of deep features. Unlike \cite{hevc_yolo,near-Lossless,chen,spatio-temporal_arrangement,chen2}, which use uniform or logarithmic quantizers to reduce the data volume, it assigns different quantization intervals to each feature channel to achieve minimal network output errors. 
	
	These codec-based approaches achieve a good balance between task efficiency and computational cost when performing machine tasks. However, these methods only consider machine vision and do not consider human vision; in other words, they do not provide a texture image that humans can view.

	\subsection{Deep Feature Compression with Leaning-based methods} 
    In \cite{end-to-end}, they propose an end-to-end feature compression system that incorporates a task-specific loss function into the training process, resulting in better analytics performance with low bit rates. In addition to providing intra-frame patterns, the system introduces inter-frame coding schemes. 
	
	Since deep features can be larger than the original image size, \cite{auto-encoder} and \cite{reduce_dimension} use an auto-encoder(AE) to reduce the feature dimension and size. Similarly, \cite{vae} introduces a Variational Auto-Encoder (VAE) model to reduce feature size while preserving task performance. 

	Like the AE,\cite{bottleFit} and \cite{rate_constrained_detection_edge} achieve feature size reduction and recovery using a front-end encoding module and a back-end decoding module, respectively. Choi \emph{et al.}\cite{scale_image_coding} also use feature reduction to decrease the amount of data, focusing on reducing feature channels rather than directly manipulating the 2-D feature size. For analysis tasks, they can use a subset of the necessary features to complete.

	End-to-end compression and feature size reduction methods reduce the required bit rate for feature transmission. However, both methods primarily focus on the analysis task and do not fully consider the reconstruction task. A hierarchical coding approach has been proposed to process both tasks jointly. Wang \emph{et al.}\cite{face} introduced a coding framework that integrates feature and texture, consisting of a base layer and an enhancement layer. In this framework, the base layer transmits the features server for the analysis task, and the enhancement layer transmits the residuals of the image reconstructed by the base layer from the original image. In \cite{human-machine}, they proposed a similar hierarchical framework using an end-to-end model that includes a viewable analysis layer and a high-quality reconstruction layer. The analysis layer is used for the analysis task, while the reconstruction layer transmits the residuals of the features used for image reconstruction.

	Hierarchical coding effectively utilizes texture and material information, enabling image reconstruction and analysis tasks to be performed simultaneously. However, as mentioned in previous methods, the residuals transmitted by the enhancement layer depend on the quality of the reconstructed image in the base layer. In features with low texture information, such as convolutional layers near the classifier, the quality of the reconstructed image is often inadequate, and in some cases, reconstruction may be impossible. Therefore, a more general layer coding framework is necessary to address these issues.

	\section{Proposed Method}
        \label{section:proposed_method}
	\begin{figure*}[htb]
		\includegraphics[scale=0.095]{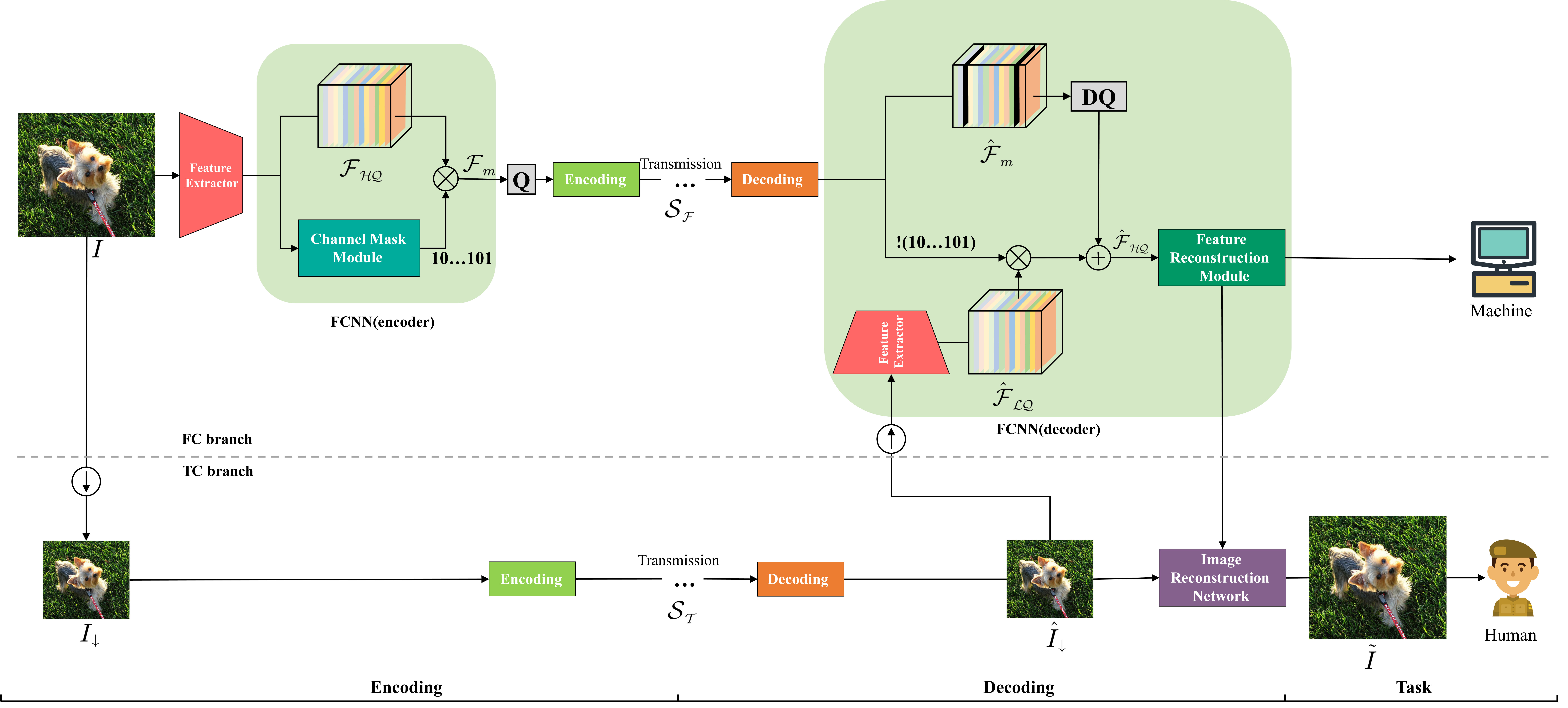}
		\caption{Overview framework of proposed method. \textcircled{$\uparrow$} is up-sample, \textcircled{$\downarrow$} is down-sample, Q is quantitation, and DQ is de-quantitation.} 
		\label{fig:encoder-decoder}
		
	\end{figure*}
	
	\subsection{Overview}

	The overall pipeline of our proposed scheme is illustrated in Fig. \ref{fig:encoder-decoder}. Our coding scheme is composed of two distinct branches. The first is the texture compression (TC) branch, and the second is the Feature Compression (FC) branch. The TC branch primarily serves humans and is dedicated to providing high-quality images to people. The FC branch serves machines and focuses on enhancing machine performance for task completion.

	The TC branch utilizes the Neural Network-based(NN) codec for image compression. Features contain highly abstract semantic information that is challenging for humans to comprehend. In conventional feature compression frameworks, it is challenging to satisfy human and machine perception requirements simultaneously. To address human visual perception, we introduce a TC branch. This branch provides preview images for humans and assists the FC branch in feature reconstruction. The workflow is as follows: at the encoding end, the source image $I$ is downsampled and fed into a NN encoder to generate the texture stream $\mathcal{S}_\mathcal{T}$, where downsampling is performed to reduce the cost of image transmission. At the decoding end, the received $\mathcal{S}_\mathcal{T}$ is decoded to obtain the down-sampled image $\hat{I}_{\downarrow}$. Then $\hat{I}_{\downarrow}$ and the decoded features of the FC branch are jointly sent to the \emph{Image Reconstruction Network} to obtain the reconstructed image $\tilde{I}$ for human.

	The FC branch is primarily used for feature compression and reconstruction. It consists of an encoder and a decoder of the \emph{Feature Compression Neural Network} (FCNN).
	At the encoding stage, the source image $I$ is fed into a feature extractor to obtain features $\mathcal{F}_{\mathcal{H}\mathcal{Q}}$, which consist of $C$ channels. However, the importance of these $C$ channels varies, and not every channel needs to be compressed and transmitted. To reduce the data volume of the features, we use a \emph{channel selection module} in the FCNN to determine which channels to keep. This module outputs the channel selection information consisting of $C$ values, where 1 indicates channel retention, and 0 indicates channel removal. The retained channels are then used to construct a feature subset $\hat{\mathcal{F}}{m}$ with $M$($M \leq C$) channels. Finally, the quantized $\hat{\mathcal{F}}_{m}$ is flattened and passed through the HEVC encoder for compression and transmission. The quantization parameters and channel selection information are also transmitted.
	At the decoding stage, the flattened features are first decoded from the bitstream $\mathcal{S}_{\mathcal{F}}$ and then rearranged to reconstruct the feature subset $\hat{\mathcal{F}}{m}$. The missing channels are filled with zeroes. However, $\hat{\mathcal{F}}{m}$ contains only partial channel information and cannot be used directly for the task at hand. To reconstruct the full feature, the downsampled image reconstruction $I{\downarrow}$ from the TC branch is upsampled and passed through the feature extractor to obtain the features $\mathcal{F}_{LQ}$. By incorporating the channel selection information, the missing channel information is extracted from $\mathcal{F}_{LQ}$ and combined with $\hat{\mathcal{F}}{m}$ to reconstruct the feature $\hat{\mathcal{F}}_{HQ}$. The process of feature recombination is explained in the next section. Finally, this feature is fed into the \emph{Feature Reconstruction Module}(FRM) of FCNN to recover the features for machine vision tasks.

	\subsection{Feature Compression Neural Network(FCNN)}
	\label{subsec:residual_encoding}

      Neural network design often involves gradually reducing feature dimensions while rapidly increasing convolutional kernel numbers to capture high-level semantic information for visual tasks. For instance, in conventional neural networks, the initial convolutional layer transforms a 3-channel RGB input image into 32 or 64 channels, halving the feature size. Consequently, the number of channels in intermediate convolutional feature maps increases more rapidly than the size reduction, leading to a substantial amount of feature data. Some researchers opt to reduce feature data by discarding channels, which may degrade task performance. To maintain task performance, researchers directly utilize the network to repair lost channels, mitigating the impact of dropped channels. However, regenerating channels filled with zeros poses a significant challenge for the network. To address this issue, we employ features extracted from texture images to substitute for the lost channels. Nevertheless, the disparity between features extracted from low-quality texture images and those from uncompressed images is considerable, leading to unsatisfactory performance improvements for visual tasks. Thus, applying an enhancement network to the reorganized features becomes necessary to enhance channel quality.
      
     Based on this observation, we propose the Feature Compression Neural Network, which consists of a \emph{Channel Selection Module} and a \emph{Feature Reconstruction Module}. In our approach, we use the channel selection module to transmit the channels that contribute significantly to the task while filtering out the channels with low contribution. This allows us to achieve compression. To ensure task performance, we incorporate both the features of low-quality images and the high-contribution features into the \emph{Feature Reconstruction Module} to recover high-quality features.

	\subsubsection{Feature extraction}
	Feature extraction serves as the initial step in feature compression. Typically, providing the input image $I$ to the pre-trained network  $TaskNet$ yields the inference results, as demonstrated below:
	\begin{equation}
		\label{eq:tasknet}
		Result = TaskNet(I, \theta),
	\end{equation}
	where $\theta$ represents the pre-trained network parameters. In the process of feature compression, the task network is divided into two networks, $TaskNet_1$ and $TaskNet_2$, with weights represented by $\theta_1$ and $\theta_2$, respectively. It is noteworthy that $\theta_1 \cup \theta_2 = \theta$. Consequently, the inference process can be redefined as follows:
	\begin{equation}
		\label{eq:tasknet1}
		\mathcal{F} = TaskNet_1(I, \theta_1)
	\end{equation}
	
	\begin{equation}
		\label{eq:tasknet2}
		Result = TaskNet_2(\mathcal{F}, \theta_2)
	\end{equation}
	The inference results of  Eq. (\ref{eq:tasknet}) and Eq. (\ref{eq:tasknet2}) are indeed equivalent. In Eq. (\ref{eq:tasknet1}), the $\mathcal{F}$ represents the extracted features obtained from the image $I$. In FCNN, we extract features $\mathcal{F}_{HQ}$ and $\mathcal{F}_{LQ}$ from the source image $I$ and the compressed low-resolution image $I_{LR}$, respectively, as follows:
	\begin{equation}
		\begin{split}
			\mathcal{F}_{HQ} &= TaskNet_1(I,\theta_1)\\
			\mathcal{F}_{LQ} &= TaskNet_1(\uparrow(I_{LR}),\theta_1),
		\end{split}
	\end{equation}
	where $\uparrow$ denotes upsampling.
	
	\begin{figure*}[htb]
		\includegraphics[scale=0.1]{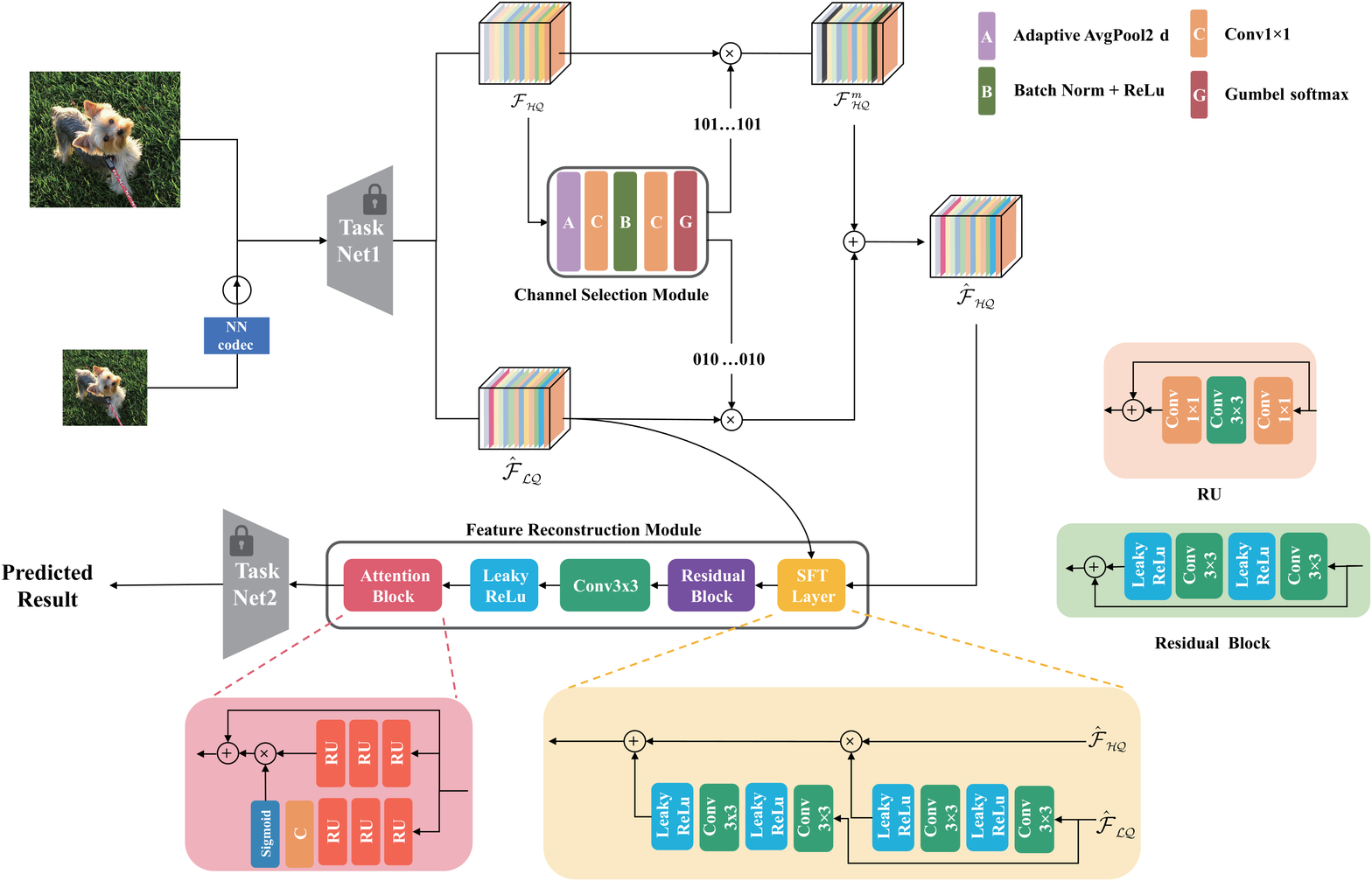}
		\caption{ Overall architecture of the Feature Compression Neural Network, including Channel Selection Module and Feature Reconstruction Module.} 
		\label{fig:FCNN}
		
	\end{figure*}

	\subsubsection{Channel Selection}
	Channel selection aims to retain task-relevant channels and filter out unimportant ones. The feature maps, denoted as $\mathcal{F}_{\mathcal{H}\mathcal{Q}}$ and $\mathcal{F}_{\mathcal{L}\mathcal{Q}}$, have dimensions of $C\times H\times W$, representing $C$ channels, each with a size of $H\times W$. Each channel contains different information (e.g., some channels may be all zeros while others have higher values overall). Additionally, different channels contribute differently to visual tasks. Treating all channels as equally important would lead to significant redundancy during compression. We employ a Channel Selection Module (CSM) to compress the features to select task-relevant channels efficiently. Inspired by the work of Xu \emph{et al.}\cite{select_frequency}, the structure of the CSM is illustrated in Fig. \ref{fig:FCNN}.
	
	The input $\mathcal{F}_{\mathcal{H}\mathcal{Q}}$ undergoes average pooling, reducing its size to $C\times 1\times 1$. It then goes through a series of operations, including a $1\times 1$ convolution, Batch Normalization, ReLU activation, and another $1\times 1$ convolution, resulting in a channel importance matrix of size $2C\times 1\times 1$. The importance matrix consists of two vectors, each with a length of $C$. The first vector represents the probabilities of selecting the current channel, while the second vector represents the probabilities of not selecting the current channel. Finally, the importance matrix is fed into the Gumbel-Softmax\cite{gumbel_softmax}, which samples channels based on their importance probabilities. The output is a binary vector of length $C$, with 1 indicating the selection of the corresponding channel and 0 indicating non-selection.
	
	For example, if $C=4$ and the output vector is $1010$, it means the first and third channels are selected. The channel selection process can be expressed as:
	\begin{equation}
		C_{one} = \text{CSM}(\mathcal{F}_{\mathcal{H}\mathcal{Q}}),
	\end{equation}
	 where $C_{one}$ is a one-hot vector of size $C\times 1$, representing the retained channels of $\mathcal{F}_{\mathcal{H}\mathcal{Q}}$. Multiplying $C_{one}$ with $\mathcal{F}_{\mathcal{H}\mathcal{Q}}$ channel-wise filters out the unimportant channels, resulting in the feature $\mathcal{F}_{m}$ that retains task-relevant channels, as shown below:
	
	\begin{equation}
		\label{eq:select_channel}
		\mathcal{F}^{i}_{m} = C^{i}_{one} \times \mathcal{F}_{\mathcal{H}\mathcal{Q}}^i, i=1,2,...,C.
	\end{equation}
	Note that during the training of the FCNN, we do not compress $\mathcal{F}_{m}$ to expedite the training process. After training the FCNN, during inference, we proceed to quantize and compress $\mathcal{F}_{m}$ as illustrated in Fig .\ref{fig:FCNN}.

	Simultaneously, by treating $C_{one}$ as a binary number and performing bitwise NOT operation on it, as shown in Eq. (\ref{eq:c_zero}), a vector of the same length, $C_{zero}$, is obtained. This vector represents the discarded channels in $\mathcal{F}_{\mathcal{H}\mathcal{Q}}$. During the feature reconstruction process, we will utilize $C_{zero}$ along with the features $\mathcal{F}_{\mathcal{L}\mathcal{Q}}$ to reconstruct the channels that were discarded in Eq. (\ref{eq:select_channel}).

	\begin{equation}
		\label{eq:c_zero}
		C^{i}_{zero} = NOT (C^{i}_{one}), i=1,2,...,C
	\end{equation}

	\subsubsection{Feature Reconstruction}
	
	Feature reconstruction aims to restore the channels that were discarded during the feature selection process. After processing through Eq. (\ref{eq:select_channel}), numerous channels in the feature $\mathcal{F}_{m}$ become zero, thereby improving the efficiency of feature compression but potentially reducing the performance in visual tasks. To enable $TaskNet_2$ to perform tasks efficiently, we utilize the features $\mathcal{F}_{\mathcal{L}\mathcal{Q}}$ to fill in the missing channels in $\mathcal{F}_{m}$. Specifically, the missing channel information in $\mathcal{F}{m}$ can be obtained by multiplying $C{zero}$ with ${F}_{\mathcal{L}\mathcal{Q}}$, as shown below:
	\begin{equation}
		\tilde{\mathcal{F}}_{m}^{i}= C_{zero}^{i} \times \mathcal{F}_{\mathcal{L}\mathcal{Q}}^{i} , i=1,2,…,C.
	\end{equation}
	Then, by adding $\mathcal{F}_{m}$ and $\tilde{\mathcal{F}}_{m}$, the reconstructed feature $\hat{\mathcal{F}}_{\mathcal{H}\mathcal{Q}}$ can be obtained:
	\begin{equation}
		\hat{\mathcal{F}}_{\mathcal{H}\mathcal{Q}} = \mathcal{F}_{m} + 	\tilde{\mathcal{F}}_{m}.
	\end{equation}

	\subsubsection{Feature Enhancement}
	
	Feature enhancement is designed to improve the quality of the reconstructed feature $\hat{\mathcal{F}}_{\mathcal{H}\mathcal{Q}}$ and further enhance task performance. The feature enhancement module consists of a Spatial feature transform (SFT)\cite{sft} Layer, a residual block, a $3\times 3$ convolution, Leaky ReLU, and an Attention Block. The SFT layer is employed to further fuse the features $\hat{\mathcal{F}}_{\mathcal{H}\mathcal{Q}}$ and $\hat{\mathcal{F}}_{\mathcal{L}\mathcal{Q}}$. Specifically, $\hat{\mathcal{F}}_{\mathcal{L}\mathcal{Q}}$ is fed into the SFT layer, which generates a set of affine parameters $(\gamma_i, \beta_i)$ through a non-linear mapping. Then, the feature $\hat{\mathcal{F}}_{\mathcal{H}\mathcal{Q}}$ undergoes channel-wise affine transformation to obtain the fused new feature $\mathcal{F}_{fusion}$, as shown below:
	\begin{equation}
		\mathcal{F}_{fusion} = \text{SFT}(\hat{\mathcal{F}}_{\mathcal{H}\mathcal{Q}} ,\hat{\mathcal{F}}_{\mathcal{L}\mathcal{Q}} ) =\gamma_i \times  \hat{\mathcal{F}}_{\mathcal{L}\mathcal{Q}}^{i} + \beta_i
	\end{equation}

	Next, the fused feature $\mathcal{F}_{fusion}$ is passed through a residual block and a $3\times3$ convolution layer, followed by Leaky ReLU activation, to further enhance feature representation. The convolutional result is then fed into an Attention Block, which weights each channel. The final enhanced feature $\hat{\mathcal{F}}$ is obtained after the channel-wise weighting. Finally, $\hat{\mathcal{F}}$ is fed into the weight-fixed $TaskNet_2$ to obtain the ultimate prediction result.

	\subsubsection{Loss Function}
	The loss function of FCNN comprises task performance loss and perceptual loss. The first part, task performance loss ($L_{T}$), aims to ensure optimal task performance with minimal rate and is defined as:
	\begin{equation}
		L_{T} = T(\hat{\mathcal{F}}) + \lambda \cdot mean(C_{one}),
	\end{equation}
	where $T(\hat{\mathcal{F}})$ represents the performance loss of $TaskNet_2$ in completing the visual task using features $\hat{\mathcal{F}}$. The $\text{mean}(C_{one})$ is a sparse loss term, indicating the number of channels to be transmitted; a higher value implies a need for more channels and a higher required bitrate. $C_{one}$ consists of C binary values, indicating selected channel positions. The calculation of $\text{mean}(C_{one})$ is given by:
	\begin{equation}
		mean(C_{one}) = \frac{1}{C} \sum_{i=1}^{C}{C_{one}^{i}}
	\end{equation}
	The parameter $\lambda$ is used to balance the trade-off between bitrate and task quality.
	
	The second part is perceptual loss ($L_{dist}$), which compensates for the performance degradation caused by discarding most channels in the features $\mathcal{F_{\mathcal{H}{\mathcal{Q}}}}$, extracted from the source image $I$ $\mathcal{F_{\mathcal{H}{\mathcal{Q}}}}$. To improve task performance, the perceptual loss is formulated as:
	\begin{equation}
		L_{dist} = \text{MSE}(\mathcal{F_{\mathcal{H}{\mathcal{Q}}}}, \hat{\mathcal{F}}).
	\end{equation}
	where $\hat{\mathcal{F}}$ is the features obtained by fusing partial channels of $\mathcal{F_{\mathcal{H}{\mathcal{Q}}}}$ and $I_{LQ}$ after processing through the feature enhancement module.
	
	Combining the task loss and perceptual loss results in the final loss used for training the network:
	\begin{equation}
		L = L_{T} + \alpha \cdot L_{dist}
	\end{equation}
	Where $\alpha$ is a weight parameter used to balance the task loss and the perceptual loss.
	\begin{figure*}[ht]
		\centering
		\includegraphics[scale=0.11]{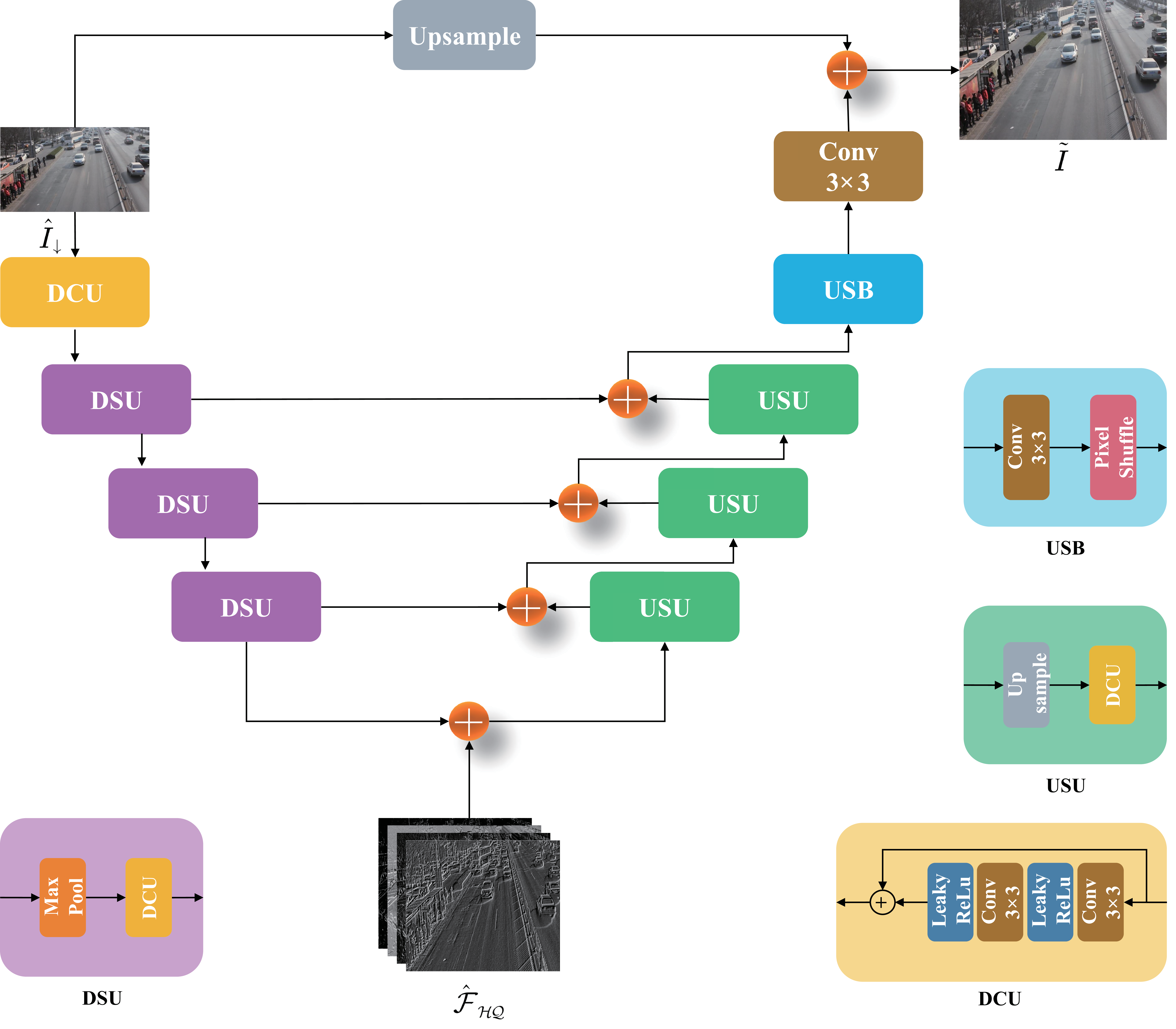}
		\caption{Detailed architecture of the proposed IRNN model including DCU, DSU, USU and USB.}
		\label{fig:IRNN}
	\end{figure*}

	\subsection{Image Reconstruction Neural Network(IRNN)}

 	Low-quality images lose semantic information due to downsampling and image compression operations. To enhance the quality of the reconstructed image, we fuse feature maps containing semantic information with the low-quality image. However, the size of the depth features and the low-quality map differ and cannot be fused directly. Moreover, the size of depth features extracted from different locations varies, posing texture and feature fusion challenges. 
    
     Through observation, we find that the size difference between different feature layers is always an integer multiple of 2. Thus, it is possible to simultaneously transform the features and texture to the same size by either up-sampling the features or down-sampling the low-quality image. Building on this observation, we propose a generalized feature fusion network named \emph{Image Reconstruction Neural Network(IRNN)} using the UNet\cite{unet} framework. The IRNN maximizes the utilization of information from both the feature and texture domains to accomplish image reconstruction. Guided by texture information, reconstructing the image from the feature domain becomes more efficient. The IRNN takes two inputs, features and texture, and outputs an enhanced image for human viewing. As Fig. \ref{fig:IRNN} shows, the IRNN is constructed by the double conv unit (DCU), the down-sampling unit (DSU), the up-sampling unit (USU), and the up-sampling block (USB), as shown in Fig. \ref{fig:IRNN}, where the kernel size of the convolution layer of each unit is $3\times3$ and stride = 1.

	More specifically, the IRNN consists of two processes: top-down and bottom-up. The top-down process serves feature extraction and fusion and is composed of DCU and DSU modules. DCU is used to extract features from textured images and consists of two $3\times 3$ convolutions and two Leaky ReLU activations. Additionally, DCU is embedded in DSU, USU, and UPU modules to adjust feature channels. DSU, in the top-down process, adjusts feature channels and sizes and is composed of a max-pooling layer and DCU. The number of DSU modules in IRNN is determined by the input feature dimensions. The bottom-up process is used to reconstruct images from features and comprises USU, USB, and a $3\times 3$ convolution. USU consists of upsampling and DCU. Upsampling employs bilinear interpolation to reduce network parameters. The features of USU processes are residual-connected to the convolution results of the same layer and serve as input to the next USU. USB consists of a $3\times 3$ convolution and PixelShuffer for feature upsampling. The convolution result from USB, after adjusting the channels with a final $3\times 3$ convolution, is added to the upsampled low-resolution image $\hat{I}_{\downarrow}$ to output the reconstructed image $\tilde{I}$. The L2 loss is adopted as the loss function in our proposed IRNN:
	\begin{equation}
		\label{loss}
		L = \frac{1}{W\times H}\sum_{x=1}^{W}\sum_{y=1}^{H}(I_{(x,y)} - \tilde{I}_{(x,y)})^2,
	\end{equation}
	where  $I$ represents the target image, $\tilde{I}$ denotes the image reconstructed by the network, and $H$ and $W$ correspond to the image's width and height, respectively. The  $x$ and $y$ represent the positions of pixels.

		\section{Experiments}
            \label{section:experiments}
		\subsection{Experimental Detail}

		In computer vision, VGGNet\cite{vgg} is widely used for feature extraction in the back-end of various visual tasks. Our work chooses the general VGG16 as the back-end for classification tasks. VGG16 consists of 13 convolutional layers, five max-pooling layers, and three fully connected layers. We disconnect the VGG16 network at the 4-th pooling layer and split it into two subnetworks, one for feature extraction and the other for classification tasks. It is worth noting that the VGG16 network used is pre-trained, and its weights are provided by PyTorch\cite{pytorch}. To maintain consistency with the pre-trained weights, the weights of the feature extraction network and the classification network are frozen.

		\subsubsection{Datasets} We selected 50,000 images from the ILSVRC2012\cite{imagenet} validation dataset to train our feature compression neural network(FCNN) and image reconstruction neural network(IRNN). The training set consists of 44,100 images, and the validation set consists of 4,900 images. We also randomly selected one image from each of the 1,000 classification categories as the test set to verify the classification performance.During training, we resized the images to $256\times256$ and applied normalization.
		
		\subsubsection{Compress Methods} In our work, we adopt classic video compression algorithms and neural network compression algorithms to compress features and images separately. Specifically, for the feature layers in FCNN, we first rearrange the filtered feature channels into a two-dimensional feature map and fill the missing parts with 0. Then, we use HM-16.12 to compress the quantized features. For the texture layer, we downsample the input image by 2 times and use the Cheng2020-Attn\cite{attention_block} model for compression for easy training.
		
		\subsubsection{Training Settings}
		To accelerate training and enable backpropagation of gradients, we simplify the feature compression process to an identity function during FCNN training, i.e., no quantization or compression of features is performed. Quantization and compression operations are only performed on features during the testing phase. For texture images, we use the Chen2022-Attn model with $\text{quality} = 4$ provided by CompressAI\cite{compressai} for compression in both the training and testing phases of FCNN and IRNN models. Additionally, the learning rate and epoch for training FCNN and IRNN are the same, which are 1e-4 and 300, respectively.

        \subsection{Effectiveness Verification for Image Classification Task}
        To verify the visual task performance of our method, we evaluate the classification performance on the ImageNet 2012 dataset. The dataset contains 1000 categories. For fairness, we randomly select one image from each category for testing. For comparison, we compress the original image using the classic video compression algorithm HEVC, denoted as Image Anchor. In addition, we extract features from the original image and then tile all feature channels into a two-dimensional grayscale image, which is then compressed using HEVC, denoted as Feature Anchor. In addition, three more popular end-to-end image compression models bmshj2018-hyperprior\cite{bmshj2018_hyperprior}, mbt2018\cite{mbt2018} and cheng-2020-anchor\cite{attention_block} are selected for comparison.

		During compression, we use the intra-configuration of HEVC to compress features and images. We selected some test points with close bitrates for model comparison. It is worth noting that features need to be quantized before HEVC can compress them. Therefore, we adopt the quantization method of Chen \emph{et al.}\cite{chen} to perform 8-bit quantization on the features, and the quantization and dequantization formulas are as follows:
		\begin{equation}
			\label{eq:quant}
			\hat{F}_{i} = round(\frac{log(F_{i} - min(F_{i}) + 1)}{max(log(F_i - min(F_i)+1))}) \cdot 255
		\end{equation}
		
		\begin{equation}
			\label{eq:dequant}
			\tilde{F}_{i} = 2^{\frac{\hat{F}_i \cdot max({log(F_{i} - min(F_{i}) + 1)})}{255}} + min(F_{i}) - 1
		\end{equation}
		Where $\hat{F}_i$ represents the quantized feature of the $i$-th channel, and $\tilde{F}_i$ denotes the dequantized $i$-th channel.

        \begin{figure}[t]
            \centering
            \includegraphics[scale=0.35]{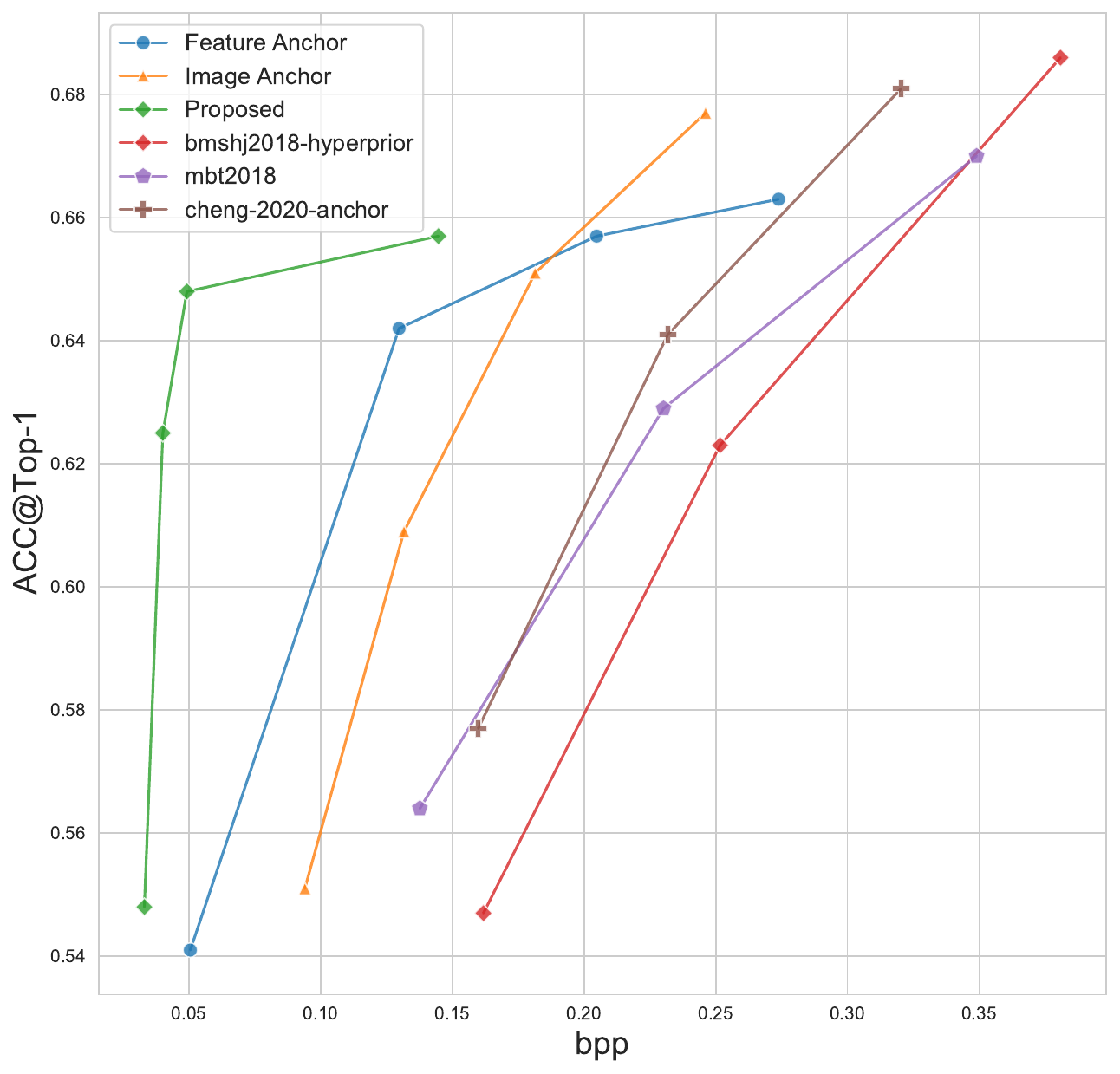}
            \vspace{-.5em}
            \caption{Rate-Accuracy curve of the proposed method and other methods for image classification.}
            \label{fig:VGG}
        \end{figure}
	 
		We use rate-accuracy curves to evaluate the performance of our proposed method. The rate is measured by bits per pixel (bpp), and the accuracy is measured by top-1 accuracy(ACC). The calculation formula of bpp is as follows:
		\begin{equation}
			\label{eq:bpp}
			bpp = \frac{total\_bits}{H\times W}
		\end{equation}
		where $W$ and $H$ represent the width and height of the original image, and $total\_bits$ denotes the total consumed bits. For our method, $total\_bits$ includes not only the bits of the feature stream and the texture stream, but also the bits consumed by the quantization parameters.
		
		The experimental results are shown in Fig. \ref{fig:VGG}. We set the weight parameters for our method to $\alpha=0.5$ and $\lambda=3$. From the figures, it can be observed that our method achieves better task performance at the same bpp under low bit rates. Compared to Image Anchor, our algorithm's BD-Accuracy can reach $-69.46\%$, saving $-69.85\%$ bits at the same ACC. Fig. \ref{fig:VGG} also indicates that using texture images alone leads to poor task performance. However, task performance significantly improves when combined with the transmission of partial channels from the feature layer. This suggests the effectiveness of our collaborative strategy between texture and feature layers. Additionally, it is noteworthy that the performance improvement of features at high bit rates is limited, possibly due to quantization errors.

        In addition to assessing the Rate-Accuracy performance, we evaluated the compression performance of our algorithm on various depth feature layers. The experimental results are summarized in Table \ref{tb:bpp}. The feature layers in the table represent features extracted from different positions of VGG16, where "Pool1" denotes a shallow feature layer with the largest data volume, and "Pool4" represents a deep feature layer with the smallest data volume. The compression rate in the table is computed by dividing the compressed data volume by the uncompressed data volume. A lower value indicates better compression performance. The calculation formula is given by:
        \begin{equation}
            \text{Compression rate} = \frac{\text{Compressed Data Volume}}{\text{Uncompressed Data Volume}} \times 100\%
        \end{equation}
        The table reveals that, under the same accuracy, our feature compression algorithm achieves a lower compression rate compared to directly using HEVC to compress features, with an average compression performance improvement of approximately 3 times. This demonstrates that our algorithm not only achieves excellent compression performance on deep feature layers but also maintains strong performance on shallow feature layers, highlighting the generality of our method.

	\begin{figure*}[htbp]
		
		\begin{minipage}{0.24\linewidth}
			\vspace{3pt}
			\centerline{\includegraphics[width=\textwidth]{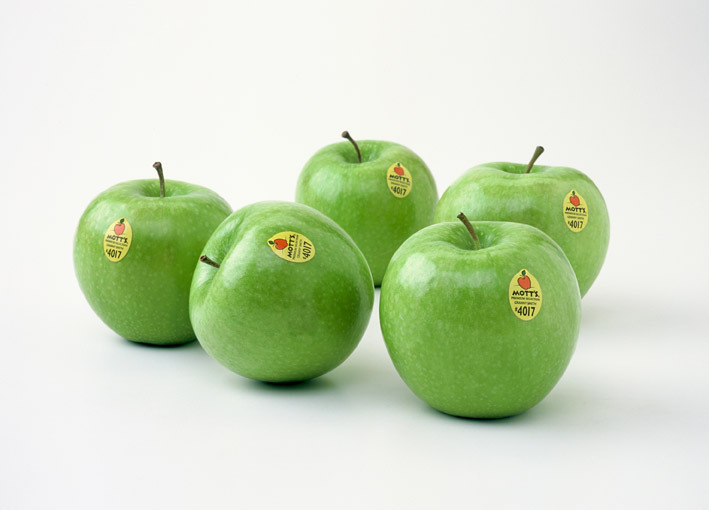}}
			\centerline{}
			\vspace{3pt}
			\centerline{\includegraphics[width=\textwidth]{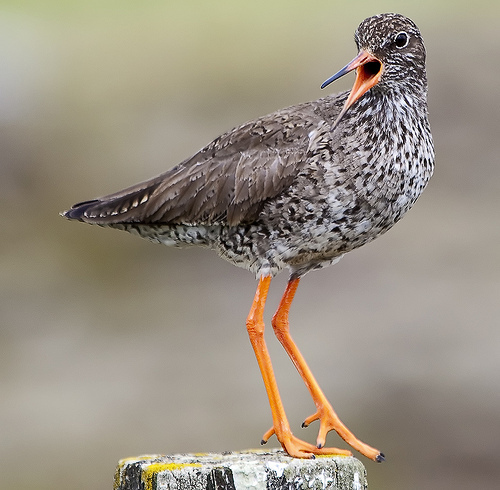}}
			\centerline{}
			\vspace{3pt}
			\centerline{\includegraphics[width=\textwidth]{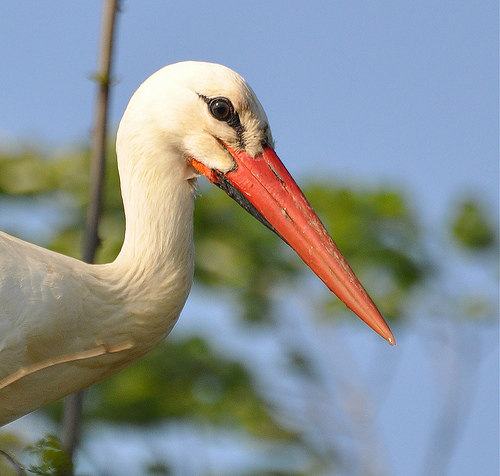}}
			\centerline{}
			\vspace{3pt}
			\centerline{(a) Ground Truth}
		\end{minipage}
		\begin{minipage}{0.24\linewidth}
			\vspace{3pt}
			\centerline{\includegraphics[width=\textwidth]{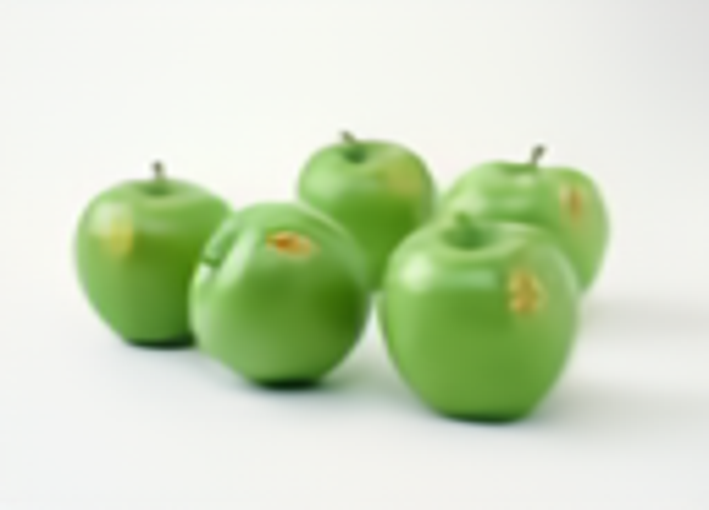}}
			\centerline{PSNR=29.322 dB}
			\vspace{3pt}
			\centerline{\includegraphics[width=\textwidth]{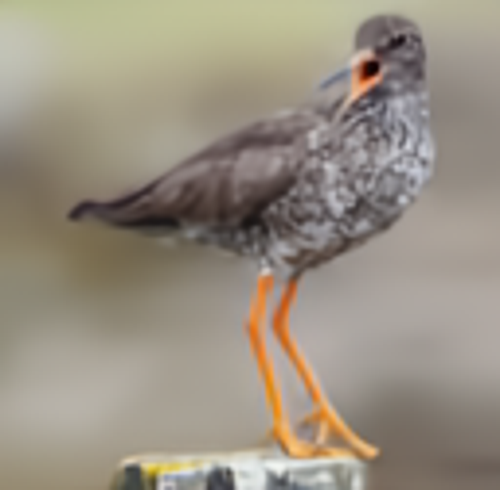}}
			\centerline{PSNR=21.542 dB}
			\vspace{3pt}
			\centerline{\includegraphics[width=\textwidth]{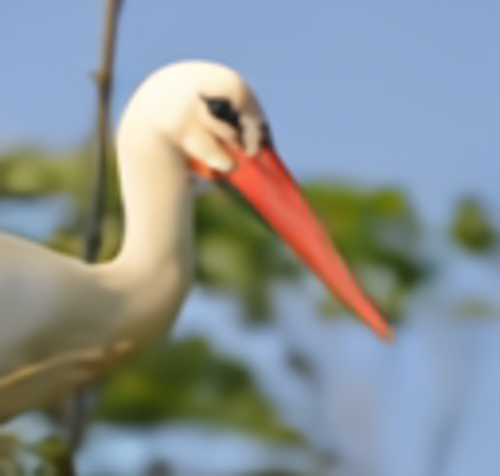}}
			\centerline{PSNR=28.447 dB}
			\vspace{3pt}
			\centerline{(b) Bicubic}
		\end{minipage}
		\begin{minipage}{0.24\linewidth}
			\vspace{3pt}
			\centerline{\includegraphics[width=\textwidth]{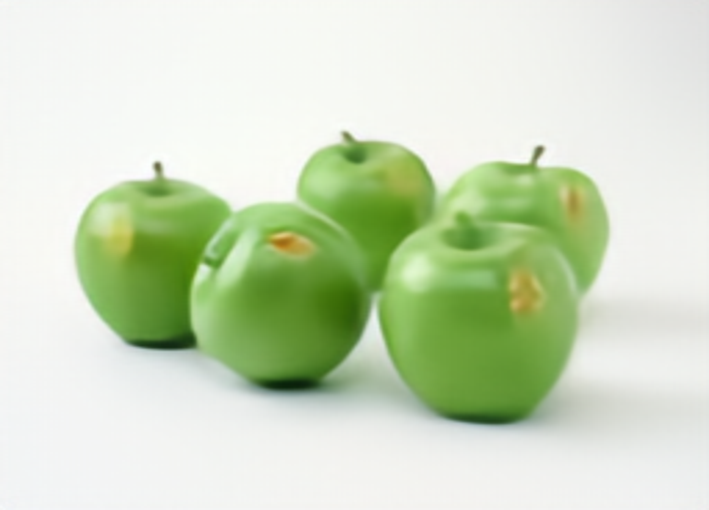}}
			\centerline{PSNR=29.996 dB}
			\vspace{3pt}
			\centerline{\includegraphics[width=\textwidth]{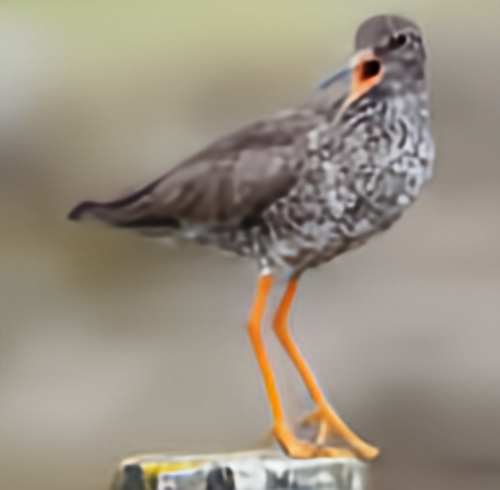}}
			\centerline{PSNR=21.780 dB}
			\vspace{3pt}
			\centerline{\includegraphics[width=\textwidth]{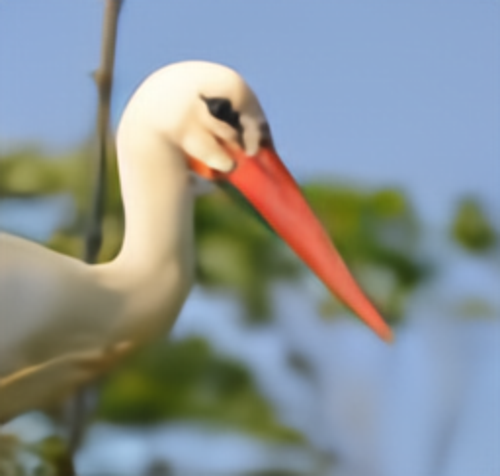}}
			\centerline{PSNR=29.095.78 dB}
			\vspace{3pt}
			\centerline{(c) Texture-SR}
		\end{minipage}
		\begin{minipage}{0.24\linewidth}
			\vspace{3pt}
			\centerline{\includegraphics[width=\textwidth]{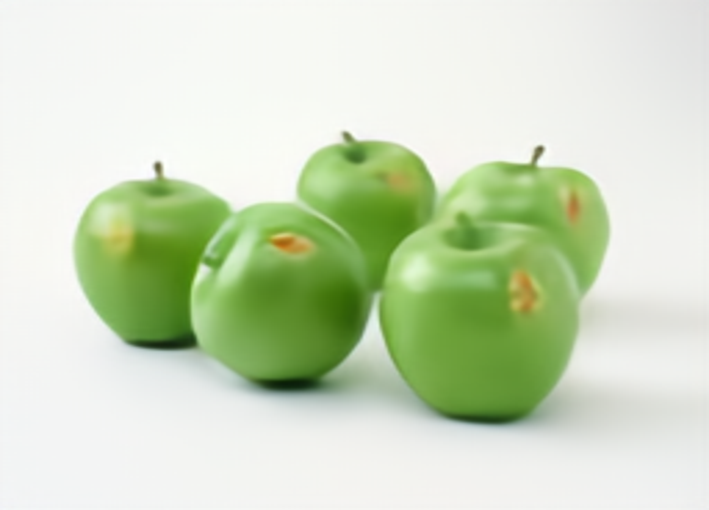}}
			\centerline{PSNR=30.171 dB}
			\vspace{3pt}
			\centerline{\includegraphics[width=\textwidth]{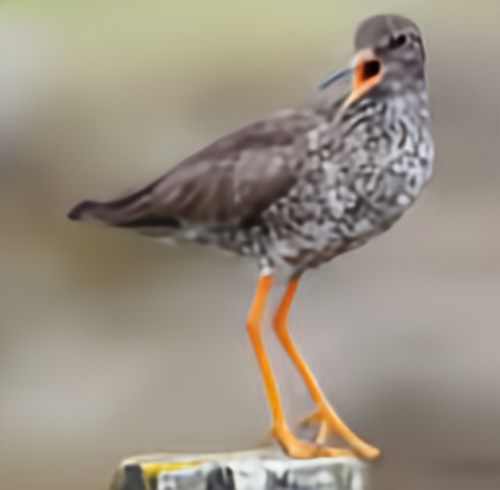}}
			\centerline{PSNR=21.802 dB}
			\vspace{3pt}
			\centerline{\includegraphics[width=\textwidth]{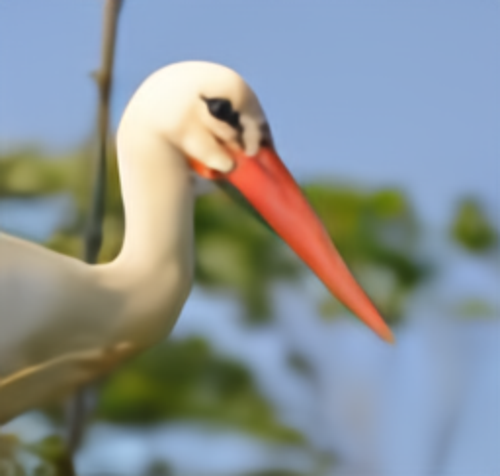}}
			\vspace{3pt}
			\centerline{PSNR=29.223 dB}
			\centerline{(d) Texture-Feature-SR}
	\end{minipage}

	\caption{Comparison of visual results. The three images have bpp values of 0.078, 0.114, and 0.114, respectively. (a) is the original uncompressed image. (b) is the image reconstructed by bicubic upsampling of the compressed texture image. (c) is the image reconstructed by IRNN with the texture image. (d) is the image reconstructed by IRNN with the texture and feature images.}
	\label{fig:preview}
	\end{figure*}
		
    \begin{table}[htbp]
        \renewcommand\arraystretch{1.2}
        \centering
        \fontsize{7}{7}\selectfont
        \caption{Rate-Accuracy performance of different feature layers}\label{tb:bpp}
    
        \begin{tabular}{c c l c c} 
            \Xhline{1.2pt}
            \hline\hline
            Feature Layer                  & Target Accuracy[\%]          & Method & BPP & Compression Rate[\%]  \\ 
            \hline
            
            \multirow{4}{*}{Pool1} & \multirow{2}{*}{$\approx$ 62} & HEVC   &0.999          &26.24        \\ 
            &                     & Proposed     &0.177            &4.66        \\ 
            \cline{2-5}
            & \multirow{2}{*}{$\approx$ 65} & HEVC   &1.495             &39.27        \\ 
            &                     & Proposed     & 0.245           &6.43        \\
            \hline
            
            \multirow{4}{*}{Pool2} & \multirow{2}{*}{$\approx$ 62} & HEVC   & 0.665          &   17.47     \\ 
            &                     & Proposed     & 0.215           &5.65        \\ 
            \cline{2-5}
            & \multirow{2}{*}{$\approx$ 65} & HEVC   & 1.075          & 28.24       \\ 
            &                     & Proposed     &0.517           & 13.59       \\
            \hline
            
            \multirow{4}{*}{Pool4} & \multirow{2}{*}{$\approx$ 62} & HEVC   &0.130           &3.41        \\ 
            &                     & Proposed     &0.040           &  1.06      \\ 
            \cline{2-5}
            & \multirow{2}{*}{$\approx$ 65} & HEVC   & 0.205          & 5.38       \\ 
            &                     & Proposed     & 0.050          & 1.30       \\
            \hline\hline
            \Xhline{1.2pt}
        \end{tabular}
    \end{table}

		\subsection{Quality Enhancement for Compressed Image}
		Guided by the texture image, our method achieves better machine vision task performance at the same bitrate. It provides humans with preview images at the decoding end to assist decision-making, meeting the "dual vision" requirements of human eye visual perception and machine vision cognition. Moreover, the features can also be applied to texture images to facilitate image reconstruction.
		
		We conducted experiments with three strategies to validate the performance of IRNN and the auxiliary role of features in image reconstruction. The first method applies Bicubic upsampling to the compressed low-resolution texture image $\hat{I}_{\downarrow}$ to restore the original resolution with a $2\times$ factor, denoted bicubic. The second method feeds $\hat{I}_{\downarrow}$  solely into the IRNN network for image reconstruction, denoted Texture-SR. The third method takes features and $\hat{I}_{\downarrow}$  as inputs for joint reconstruction by IRNN, denoted Texture-Feature-SR. The experimental setup remains the same as the previous section, with 1000 images used for evaluation. The feature's QP and texture quality are set to 11 and 4, respectively. 
		
\begin{figure*}[htbp] 
			\centering
			\subfloat[]{
				\includegraphics[height=4cm, width=4cm]{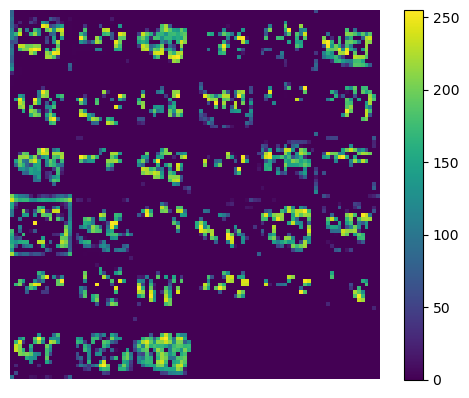}
			}
			\subfloat[]{
				\includegraphics[height=4cm, width=4cm]{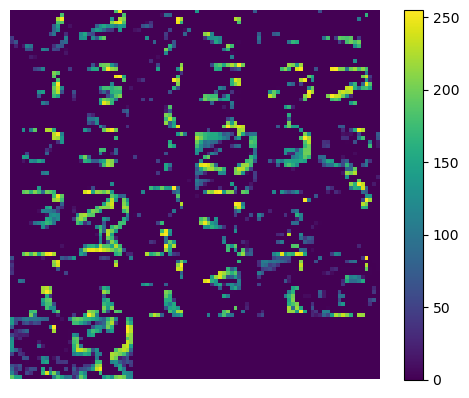}
			}
			\subfloat[]{
				\includegraphics[height=4cm, width=4cm]{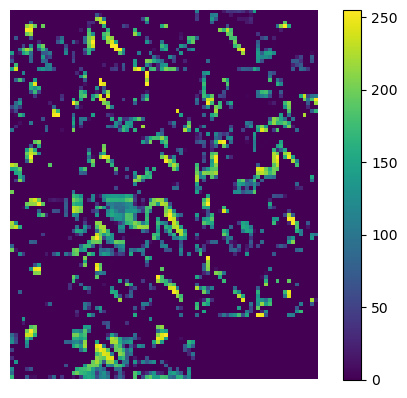}
			}
			\vspace{3pt}
			\caption{Visualization of feature channels. (a), (b), and (c) correspond to the feature maps of the three images in Figure 5, respectively. These feature maps are used for image reconstruction together with the texture images.}
			\label{fig:feature}
	\end{figure*}

		The experimental results are as follows: The average PSNR for the first method is 23.254dB, for the second method is 23.949dB, and for the third method is 24.009dB. Comparing the results of the first and second methods, our IRNN reconstruction network improves performance by 0.695dB. 
		A comparison between Bicubic and Texture-SR demonstrates that our IRNN reconstruction network can improve performance by 0.695 dB. Furthermore, a comparison between Methods Texture-SR and Texture-Feature-SR reveals that features can guide image reconstruction. As shown in Fig. \ref{fig:preview}, the target of the preview image reconstructed by the joint features and texture is slightly clearer. 
		However, the improvement is not significant compared to Texture-SR. This could be attributed to our feature selection strategy, as depicted in Fig. \ref{fig:feature}. Since the feature layer close to the classifier inherently contains less texture information, and the feature selection module tends to prioritize features relevant to machine vision tasks,  the improvement is not substantial.

        \subsection{Ablation Study}

        \begin{table}[htbp]
            \renewcommand\arraystretch{1.2}
            \centering
            \fontsize{7}{7}\selectfont
            \caption{Ablation experiment setup}\label{tb:ablation}
            
            \begin{tabular}{lcc}
            \Xhline{1.2pt}
            \hline\hline
            
                Method & Texture & FRM \\
                \midrule
                1. FCNN w/o texture and FRM & \ding{55} & \ding{55} \\
                2. FCNN w/o FRM  & \checkmark & \ding{55}  \\
                3. FCNN w/o Texture & \ding{55} & \checkmark  \\
                4. FCNN & \checkmark & \checkmark \\
                \hline\hline
                \Xhline{1.2pt}
            \end{tabular}
        \end{table}

        \begin{figure}[htbp]
            \centering
            \includegraphics[scale=0.35]{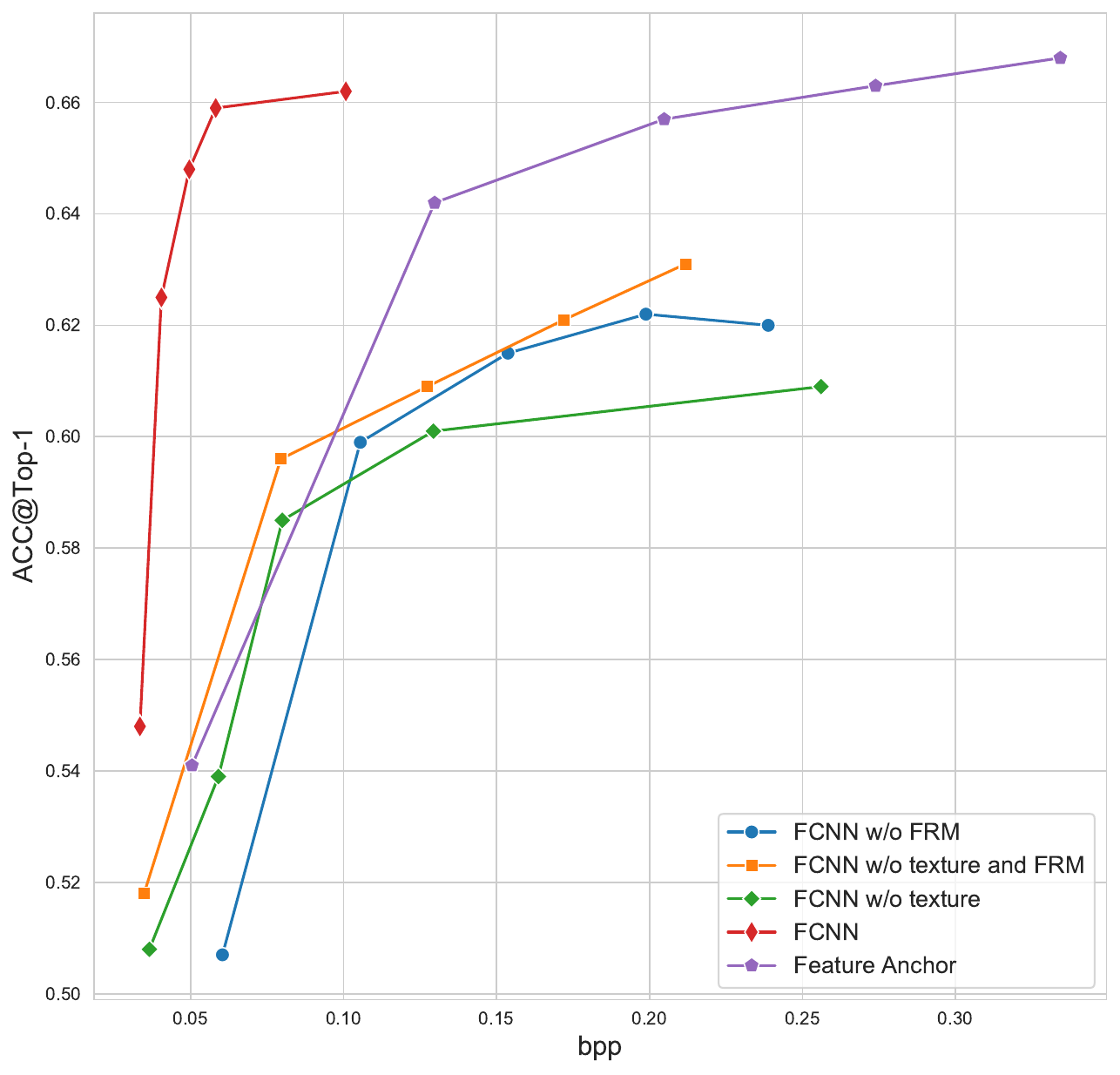}
            \vspace{-.5em}
            \caption{ Rate-Accuracy  results of image classification for different comparison methods}
            \label{fig:ablation}
		\end{figure}

        In order to assess the impact of two strategies on detection performance, we conducted ablation experiments. The strategies in question are direct channel discarding and using low-quality features to fill the missing channels. The experimental setups are detailed in Table \ref{tb:ablation}. The "Texture" parameter indicates whether low-quality features extracted from the texture are used to replace the lost channels. When "Texture" is enabled, it means that features $\hat{\mathcal{F}}_{HQ}$ from Fig. \ref{fig:FCNN} are utilized for subsequent training; otherwise, the network is trained using $\mathcal{F}^{m}_{HQ}$. The "FRM" parameter indicates whether the FRM is employed to reconstruct the features $\hat{\mathcal{F}}_{HQ}$ or $\mathcal{F}^{m}_{HQ}$ for enhancement.

        The experimental results, presented in Fig. \ref{fig:ablation}, reveal comparisons between Method 1 and Method 2. It is observed that the classification performance when directly discarding feature channels and when utilizing low-quality features to fill in the lost channels is comparable, leading to a degradation in classification performance. Moreover, upon closer examination, filling the lost channels in Method 1 directly with low-quality feature channels results in a slight decrease in classification performance, while replacing the filled low-quality graphs in Method 2 with zeros maintains their classification performance. These findings suggest that the gap between low-quality and high-quality features is too significant to directly fill in the missing feature channels, thus limiting the potential improvement in visual task performance.
        
        Additionally, a comparison between Method 1 and Method 3 demonstrates that, with the same QP compression and the assistance of the FRM module, Method 1 can further reduce the amount of data for the features—i.e., it can discard more channels—but at the cost of decreased classification performance.
        
        Furthermore, contrasting Method 2 with Method 4 reveals that, with the assistance of the FRM module, it can further reduce the amount of data for the features, and the performance of the visual task remains unchanged. This underscores the efficient ability of FRM to reconstruct and enhance features.

		\section{Conclusion}
            \label{section:conclusion}
		In this paper, we propose a texture-guided deep feature compression method. The proposed method consists of a two-layer encoding structure: the feature layer and the texture layer. The feature layer is designed to provide efficient deep features for machine vision tasks, while the texture layer aims to provide humans with preview images for decision-making assistance. Specifically, the feature layer consists of a channel selection module and a feature reconstruction module. During encoding, the channel selection module directly discards task-irrelevant channels to save bits, and the missing feature channels are recovered using the texture image during decoding to ensure the quality of machine vision tasks. The texture layer consists of an image reconstruction network that utilizes the features from the feature layer to guide the image reconstruction process.
		
		Experimental results demonstrate that the proposed method can perform better vision tasks at lower bitrate consumption. Under texture guidance, the proposed method can effectively remove channel redundancy in features, thereby reducing the amount of feature data. Meanwhile, the features can also assist image reconstruction, which addresses the needs of both human and machine vision. 
		
		In the future, we will explore using end-to-end networks to replace traditional video codecs for generating bitstreams. This will enable end-to-end training and reduce the impact of different quantization parameters on tasks.

\end{document}